\documentclass[conference]{IEEEtran}

\usepackage{cite}
\usepackage{amsmath,amssymb,amsfonts}
\usepackage{algorithmic}
\usepackage{graphicx}
\usepackage{textcomp}
\usepackage{xcolor}
\usepackage{booktabs}
\usepackage{multirow}
\usepackage{dblfloatfix}
\usepackage{tabularx}
\usepackage{hyperref}

\hypersetup{
    colorlinks=true,
    linkcolor=blue,
    citecolor=blue,
    urlcolor=blue
}
\begin{document}

\title{Denoise X: Self-Supervised U-Net for Quantum Mottle Suppression in Chest Radiography}

\author{
\begin{tabular}{cc}
\begin{tabular}{c}
\textbf{Parth Giramkar}\\
Department of Artificial Intelligence and Data Science\\
Dr. D. Y. Patil Institute of Technology\\
Pune, Maharashtra 411018, India\\
imparthgiramkar@gmail.com
\end{tabular}
&
\begin{tabular}{c}
\textbf{Aryan Patle}\\
Department of Artificial Intelligence and Data Science\\
Dr. D. Y. Patil Institute of Technology\\
Pune, Maharashtra 411018, India\\
aryanpatle031@gmail.com
\end{tabular}
\\[12mm]

\multicolumn{2}{c}{
\begin{tabular}{c}
\textbf{Vanita Kshirsagar}\\
Department of Artificial Intelligence and Data Science\\
Dr. D. Y. Patil Institute of Technology\\
Pune, Maharashtra 411018, India\\
vanita.kshirsagar@gmail.com
\end{tabular}
}
\end{tabular}
}

\maketitle

\begin{abstract}
Deep learning approaches have demonstrated significant potential for medical image restoration; however, their clinical application is often constrained by the requirement for large-scale paired training data under supervised learning. In particular, acquiring perfectly registered clean--noisy radiograph pairs \textit{in vivo} is impractical due to patient motion, physiological variations, and radiation-dose constraints governed by the ALARA (As Low As Reasonably Achievable) principle. To address this limitation, a self-supervised Noise2Noise learning paradigm is employed, enabling training using independently corrupted observations without requiring paired clean images while preserving clinically important anatomical structures. Based on this approach, Denoise X is proposed as a lightweight self-supervised U-Net-based model with 7.79 million parameters for suppressing quantum mottle and electronic readout noise in low-dose chest radiographs. The model is trained entirely on the NIH ChestX-ray14 dataset containing 112,120 images using a stochastic Poisson--Gaussian noise simulator to generate realistic radiographic degradations. Zero-shot evaluation on the unseen Montgomery County dataset achieved a PSNR of 36.09~dB and an SSIM of 0.8806, substantially outperforming the Bilateral filter in structural similarity while preserving high-frequency anatomical details. Using patch-wise inference, the proposed model processes native-resolution multi-megapixel chest radiographs in approximately 2.2~s per image on a Tesla P100 GPU. Cross-domain evaluation on an unseen Shenzhen dataset image further achieved a PSNR of 36.44~dB and an SSIM of 0.9099, demonstrating promising generalization across imaging distributions. For resource-constrained deployment, the trained model was converted into a lightweight 31.16~MB ONNX model for CPU inference, with adaptive preprocessing enabling efficient processing of high-resolution non-DICOM images.
\end{abstract}

\vspace{8pt}


\vspace{2pt}

\section{Introduction}

The use of chest X-rays is one of the most common forms of medical imaging worldwide and serves as a first-line diagnostic tool for numerous cardiopulmonary disorders~\cite{wang2017chestxray8}. To ensure patient safety during examinations, radiation exposure should be minimized according to the As Low As Reasonably Achievable (ALARA) principle, which implies minimal exposure of patients to ionizing radiation~\cite{bushberg2011essential}. However, any decrease in photon flux necessarily reduces the signal-to-noise ratio of the detector, leading to increased quantum mottle (photon shot noise) and electronic noise that can mask high-frequency anatomical structures~\cite{bushberg2011essential}. Therefore, the development of advanced deep learning-based denoising techniques is crucial.

Traditional spatial filters such as Bilateral~\cite{tomasi1998bilateral} and Median filters are commonly used to address this problem. Despite being computationally inexpensive, these methods rely on local neighborhood statistics and may suppress image noise together with diagnostically important structural edges. In contrast, supervised deep learning frameworks~\cite{zhang2017beyond,ronneberger2015unet} have demonstrated promising results in medical image restoration. However, these frameworks face a significant data bottleneck, as obtaining precisely aligned pairs of noisy and clean radiographs is difficult in in-vivo environments due to patient motion and radiation exposure constraints.

Recent developments in deep learning have led to highly effective frameworks such as vision transformers~\cite{zamir2022restormer} and advanced convolutional networks~\cite{chen2022simple}, which demonstrate excellent performance in image restoration. However, their increased computational complexity may limit their use in resource-constrained healthcare environments. Processing native-resolution, multi-megapixel clinical chest X-rays through such architectures can require substantial computational resources and hardware acceleration.

The paired-data limitation can be addressed using the Noise2Noise technique~\cite{lehtinen2018noise2noise}. Given the independence and zero-mean assumption of the radiographic noise, minimizing the Noise2Noise loss can recover the underlying clean signal without requiring clean ground-truth images. Building upon this principle, \textit{Denoise X}is proposed as a lightweight U-Net architecture trained on the NIH ChestX-ray14 dataset~\cite{wang2017chestxray8} using a stochastic Poisson--Gaussian noise model~\cite{foi2008practical}. Additionally, patch-wise inference is employed for native-resolution chest radiography while maintaining compatibility with standard CPU-based deployment.

The main contributions of this work are as follows:

\begin{itemize}
    \item \textbf{Self-Supervised Architecture:} A lightweight, self-supervised Noise2Noise U-Net architecture with 7.79 million parameters is introduced, trained using a stochastic Poisson--Gaussian simulator without requiring paired clean ground-truth data.

    \item \textbf{Edge-Optimized Inference Framework:} A patch-wise inference framework optimized for ONNX CPU runtimes is presented, enabling native-resolution inference of multi-megapixel chest X-rays in resource-constrained environments.

    \item \textbf{Zero-Shot Evaluation:} Zero-shot evaluation on the unseen Montgomery County dataset demonstrates that the proposed method achieves 36.09 dB PSNR and 0.8806 SSIM, where the proposed method achieves
36.09~dB PSNR and 0.8806 SSIM, substantially outperforming conventional
spatial filtering techniques.

    \item \textbf{Cross-Domain Generalization:} Generalization to a different imaging domain is demonstrated through quantitative and qualitative evaluation on an unseen chest X-ray from the Shenzhen dataset.
\end{itemize}

\section{Related Work}

\subsection{Traditional Methods}
Traditional spatial and frequency domain filtering has been used historically for the elimination of quantum mottle and electronic noise in digital radiography. Filters like Gaussian smoothing, Median filtering, and the Bilateral filter, which has been presented by Tomasi and Manduchi \cite{tomasi1998bilateral}, are common tools used in clinical radiographic software applications due to their simplicity and minimal computational requirements. Of these, the Bilateral filter is especially useful for edge preservation through local pixel weighting depending on spatial closeness and similarity of intensity values. The limitation of these techniques lies in their reliance on local statistics for noise suppression. Since low-dose chest radiographs often have high noise variance and anatomical edges characterized by low contrast, it leads to unwanted smoothing of high-frequency anatomical structures alongside the noise.

\subsection{Supervised Image Restoration Using CNN}

In recent years, Convolutional Neural Networks (CNNs) have established strong benchmarks for image restoration tasks. Representative architectures include DnCNN~\cite{zhang2017beyond}, proposed by Zhang \textit{et al.}, which employs residual learning for effective image denoising, and U-Net~\cite{ronneberger2015unet}, originally proposed by Ronneberger \textit{et al.} for biomedical image segmentation and subsequently adopted for various image restoration tasks. Both architectures have demonstrated strong capabilities in learning mappings between degraded and restored images. However, despite their effectiveness, their application to medical radiography is fundamentally constrained by the limited availability of paired training data. Supervised learning models typically require large-scale datasets containing precisely aligned noisy and clean image pairs. Although synthetic noise can be introduced into clean images, obtaining real-world paired low-dose and clean radiographs is challenging because repeated acquisitions may differ due to patient motion, physiological variations, and changes in imaging conditions. Furthermore, repeated radiation exposure solely for obtaining paired training data is undesirable from an ethical and clinical perspective..

\subsection{Self-Supervised Image Restoration}
One solution to the problem above is to use self-supervised image restoration approaches. Lehtinen \textit{et al.} \cite{lehtinen2018noise2noise} proposed a Noise2Noise approach in which a neural network learns image restoration from independently corrupted samples of the same scene. Under the assumption of having zero-mean and statistically independent corrupting noise over observations, the optimization procedure with the $L_2$ loss leads to the expected result of the image without any noise. Following the idea described above, Noise2Void approach suggested by Krull \textit{et al.} \cite{krull2019noise2void} extended it by allowing training from one single image with noise, though usually sacrificing restoration quality. Such learning approaches are especially useful in case of medical imaging since getting such a clean image is usually impossible.

\subsection{Modern Restoration Architectures}
New developments in image restoration tasks have led to the creation of increasingly complex deep learning architectures with the ability to achieve state-of-the-art results. For instance, Restormer \cite{zamir2022restormer}, an architecture proposed by Zamir \textit{et al.}, uses efficient multi-head self-attention models in a transformer architecture for high-quality image restoration, whereas NAFNet \cite{chen2022simple}, proposed by Chen \textit{et al.}, is a computationally efficient architecture which uses convolutional layers and does not use any nonlinear activation functions yet still retains high restoration capabilities. While these architectures provide impressive results on natural image benchmarks, the computational burden of these architectures makes their usage challenging in resource-limited clinical settings. As medical radiographs are captured at native multi-megapixel resolutions, it is computationally demanding to process these high-resolution images using transformer architectures and very deep convolutional networks.

\subsection{Research Gap}
Although significant advances have been made in image restoration tasks through existing methods, they either rely on paired supervised learning or are highly computationally intensive making their practical use in a clinical setting difficult. Due to these shortcomings, \textit{Denoise X} is introduced as a light-weight self-supervised U-Net architecture that integrates the Noise2Noise learning philosophy with an efficient patch-based inference mechanism for restoration of native-resolution chest radiographs. The proposed model is designed to preserve clinically significant anatomical details while ensuring practical implementation on low-end hardware systems.

\section{Proposed Methodology}
This section describes the methodology for the development and deployment of the Denoise X framework. Overall, the methodology consists of three main steps: data and noise simulation, self-supervised denoising, and high-resolution inference and deployment. First, the NIH ChestX-ray14 database is used to generate independent noisy observations via a stochastic Poisson-Gaussian noise model, thus enabling realistic quantum mottle and electronic noise simulation without paired clinical acquisitions. The noisy observations are then used in a Noise2Noise paradigm to train a lightweight U-Net regularized by an $L_1$ loss to recover the underlying clean image from the noisy input. Finally, the trained model is deployed using a patch-wise inference strategy to enable high-resolution radiograph processing and accelerated by ONNX for CPU inference in low-resource settings. The following subsections describe the details of each step.
\begin{figure*}[!t]
    \centering
    \includegraphics[width=\textwidth]{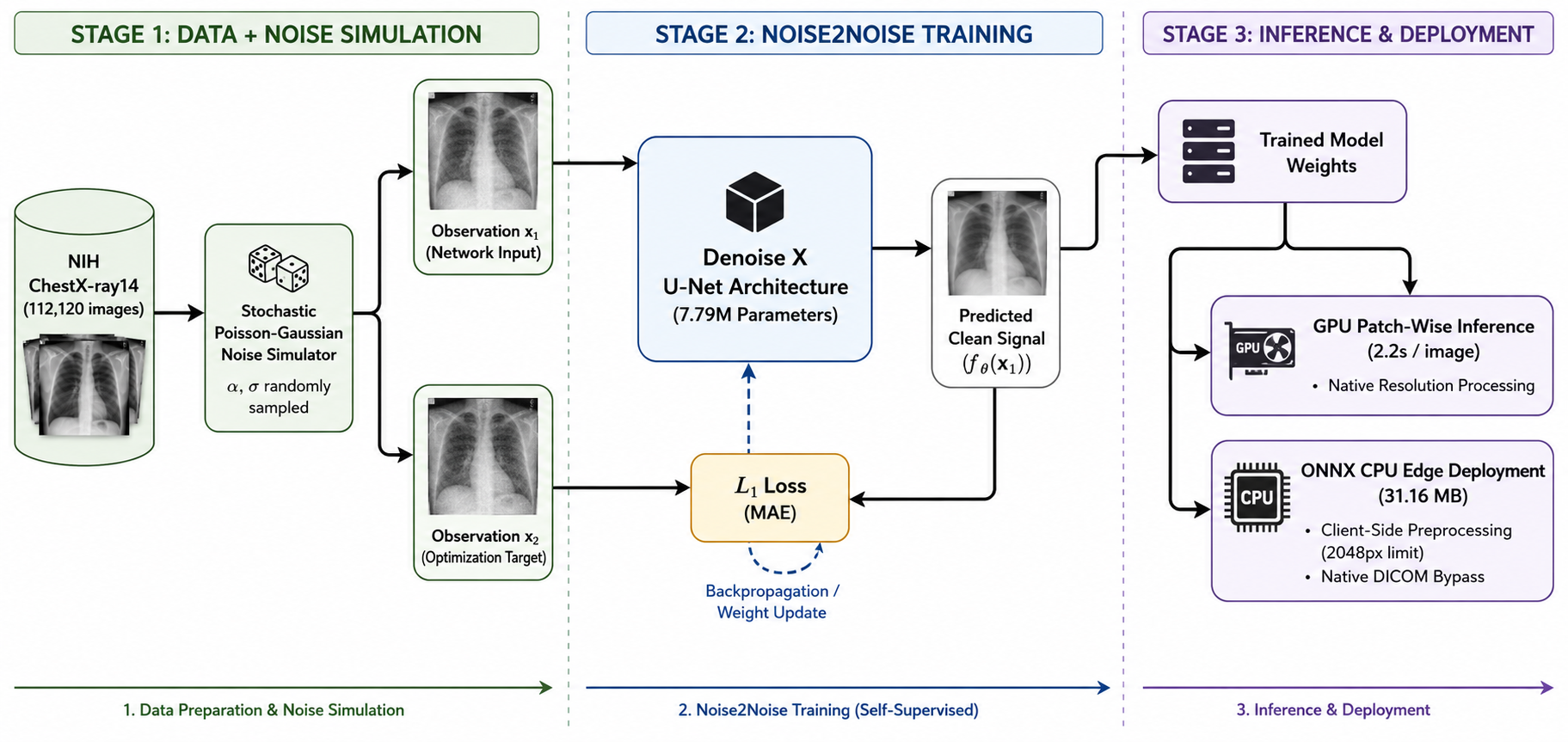}
    \caption{End-to-end workflow of the proposed \textit{Denoise X} framework. The pipeline illustrates stochastic Poisson--Gaussian noise simulation for Noise2Noise training, self-supervised U-Net optimization, and deployment using GPU-based patch-wise inference and ONNX-based CPU inference.}
    \label{fig:pipeline_flowchart}
\end{figure*}

\subsection{Modeling of Physical Noise}
The capturing of digital chest x-rays basically depends on the measurement of X-ray photons. In order to ensure safety as per the clinical radiation protocols, the amount of photons is limited. The degradation that results due to the low dose radiography process is basically Quantum Mottle. This is because it originates due to random photon arrival and hence follows the Poisson statistics. This degradation is further affected by the Gaussian distribution of electronic readout noise due to the digital detector device.

To train the restoration network without the use of clinically unavailable clean pairs, a Poisson--Gaussian compound noise model is constructed for the NIH ChestX-ray14 dataset. Given a normalized x-ray $I \in [0,1]$, the noisy observation $I_{\mathrm{noisy}}$ is modeled as

\begin{equation}
I_{\mathrm{noisy}}
=
\frac{\operatorname{Poisson}(\alpha I)}{\alpha}
+
\mathcal{N}(0,\sigma^{2})
\label{eq:poisson_gaussian}
\end{equation}

where $I$ stands for the clean anatomical image and $\alpha$ denotes the scaling parameter that controls the number of simulated photons (i.e., the simulated dose level, which is proportional to $\alpha$). The function $\mathcal{N}(0, \sigma^2)$ stands for a Gaussian distribution with the mean value of 0 and variance of $\sigma^2$ that characterizes the additive noise of electronic sensors. During the training procedure, both $\alpha$ and $\sigma$ parameters are selected randomly for each input image , thus exposing the network to a large set of diverse simulated acquisition conditions. The continuous changes in the noise properties help prevent the framework from overfitting to any specific hardware, thereby enhancing its generalization to various dose levels and digital detector properties.

\subsection{Self-Supervised Noise2Noise Learning Framework}
In traditional image denoising, pairs of perfectly aligned noisy and noise-free reference images are used for supervised training. In order to solve this problem, the self-supervised Noise2Noise framework is used. Unlike supervised denoising, where the ground truth targets are required for training, the network is trained using two independently corrupted copies of the same image.

Let $y$ be the latent image without any form of noise, and let $x_1$ and $x_2$ be two noisy observations generated using the stochastic Poisson-Gaussian model for noise generation as follows:
\begin{equation}
\begin{aligned}
x_1 &= y + n_1,\\
x_2 &= y + n_2.
\end{aligned}
\label{eq:noisy_observations}
\end{equation}
where $n_1$ and $n_2$ represent independent observations of the compound noise process. In the training phase, the input to the network is $x_1$, while $x_2$ is used as the target observation.

The network $f_\theta(\cdot)$, characterized by learnable weights $\theta$ , is optimized by minimizing the mean absolute error between its prediction and the second noisy observation : 

\begin{equation}
\mathcal{L}(\theta)
=
\mathbb{E}
\left[
\left\|
f_\theta(x_1)-x_2
\right\|_1
\right].
\end{equation}

where $\mathcal{L}(\theta)$ denotes the objective loss function, and $\mathbb{E}[\cdot]$ represents the statistical expectation operator over the training data distribution.
Given the assumption that the corruption is independent of each other and that its mean is zero, the expected value of the target corrupted with noise will be equal to the expected value of the source. Hence, by optimizing for the Noise2Noise loss function, we converge to the same expected value as with traditional supervised learning.

This property of the loss function makes it very suitable for medical imaging applications, where it is impossible from both physical and ethical points of view to have pairs of clean and corrupted radiographs. The learning framework is realized using a simple U-Net architecture aimed at preserving the anatomical features.

\subsection{Denoise X Network Architecture}
To infer the underlying anatomical signal from the observation, a lightweight fully convolutional U-Net architecture is used. The network uses a single channel grayscale tensor of spatial size $256 \times 256$ as the input and generates a restored tensor of spatial size $256 \times 256$.

The network architecture is symmetric in nature with an encoder-decoder structure:
\begin{itemize}
    \item \textbf{Encoder:} The architecture gradually down-samples the spatial resolution while simultaneously increasing the depth of the feature channels and capturing the high-level structure representation of the anatomy.
    \item \textbf{Bottleneck:} This acts as a link between the encoder and decoder pathways which performs operation on the most down-sampled feature maps.
    \item \textbf{Decoder:} Gradually up-samples the feature representation to reconstruct the spatial resolution.
    \item \textbf{Skip Connections:} In order to avoid the loss of the finer spatial resolution (fine anatomical structures like vascular structures in lung, rib margins, soft tissue interfaces) due to down-sampling, skip connections are made.
\end{itemize}

This kind of architecture allows the network to exploit both localized high-frequency information as well as anatomical context information. The suggested architecture involves around \textbf{7.79 million trainable parameters}, which offer enough representational power to restore anatomical structures and yet remain lightweight to be deployable both on GPUs and CPUs. Contrary to other medical image restoration methods using transformers, the suggested architecture consists only of convolutional layers and has lower computational complexity.

\subsection{Patch-Based High-Resolution Inference}
One of the problems associated with medical image restoration is the difference between fixed network input dimensions and resolution of native clinical radiographs. In many practical cases, downsampling of high-resolution images is done because of limited memory, thus losing important high-frequency information, quantum mottle, and anatomical edges.

To perform inference on multi-megapixel native chest radiographs without global downsampling, a high-resolution patch-wise inference strategy is employed. The inference process runs in sequence according to the following steps:
\begin{enumerate}
    \item \textbf{Reflect Padding:} The native full-resolution image gets dynamically padded with reflect padding such that both dimensions get perfectly divisible by the $256 \times 256$ window size, avoiding border artifacts due to convolution.
    \item \textbf{Patch Extraction:} The padded image is then extracted in non-overlapping $256 \times 256$ spatial patches with a stride of 256 pixels in both dimensions.
    \item \textbf{Batch Inference:} These patches are grouped in one tensor for batch inference in the U-Net model.
    \item \textbf{Patch Reconstruction:} The reconstructed patches are then concatenated to form a continuous high-resolution matrix, and the reflection padding applied at the beginning is trimmed, resulting in the final native size of the image.
\end{enumerate}

\begin{figure*}[!t]
    \centering
    \includegraphics[width=\textwidth]{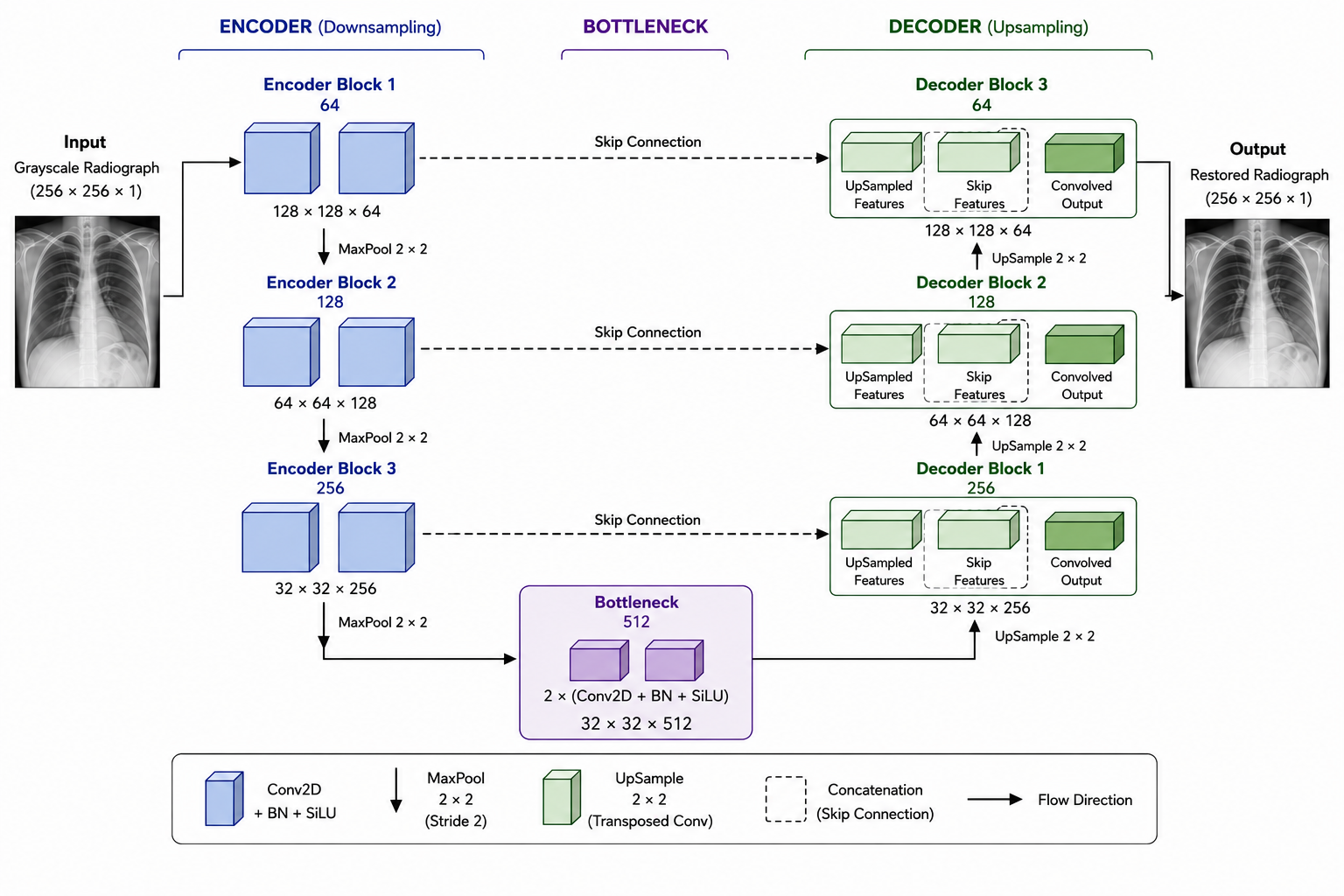}
    \caption{Architecture of the proposed \textit{Denoise X} U-Net. The network follows a symmetric encoder--decoder architecture with skip connections that preserve fine-grained anatomical details during image restoration.}
    \label{fig:unet_architecture}
\end{figure*}

This technique allows the network to focus on restoring the structure of the image at the native resolution, while still being able to infer on any size of the radiograph without changing the training of the network.

\subsection{Edge Deployment Pipeline}

To ensure the clinical feasibility of the proposed framework, the model trained in TensorFlow is exported to the Open Neural Network Exchange (ONNX) format. The exported ONNX model takes up a mere \textbf{31.16~MB}, which allows for CPU inference without any need for the TensorFlow library.

In addition, to speed up the inference for mobile and web applications, a preprocessing pipeline for the client-side HTML5 Canvas input is implemented. The preprocessing ensures that extreme resolution images have their dimensions limited to 2048~pixels (except DICOM images, which are passed natively) while maintaining the original aspect ratio. Thus, the number of inference patches is considerably reduced, allowing native CPU inference of the patch-wise restoration algorithm in about \textbf{3 to 5~seconds} per image without any dedicated GPU hardware.

\begin{table}[!t]
\caption{Architectural Details of the Proposed Denoise X U-Net (7.79M Parameters)}
\label{tab:architecture}
\centering
\footnotesize
\renewcommand{\arraystretch}{1.15}

\begin{tabularx}{\columnwidth}{@{}lXc@{}}
\toprule
\textbf{Network Component} &
\textbf{Primary Operations} &
\textbf{Output Dimension} \\
\midrule

Input &
Grayscale Radiograph &
$256\times256\times1$ \\

Encoder Block 1 &
Downsample + Conv2D &
$128\times128\times64$ \\

Encoder Block 2 &
Downsample + Conv2D &
$64\times64\times128$ \\

Encoder Block 3 &
Downsample + Conv2D &
$32\times32\times256$ \\

Bottleneck &
$2\times$(Conv2D + BN + SiLU) &
$32\times32\times512$ \\

Decoder Block 1 &
Upsample + Concat + Conv2D &
$64\times64\times256$ \\

Decoder Block 2 &
Upsample + Concat + Conv2D &
$128\times128\times128$ \\

Decoder Block 3 &
Upsample + Concat + Conv2D &
$256\times256\times64$ \\

Output Block &
$1\times1$ Conv2D (Linear) &
$256\times256\times1$ \\

\bottomrule
\end{tabularx}
\end{table}

\section{Experiments and Results}

\subsection{Datasets and Performance Metrics}

Experiments to evaluate the proposed \textit{Denoise X} framework were conducted on three publicly available chest radiograph datasets:

\begin{enumerate}
    \item \textbf{Training (NIH ChestX-ray14 \cite{wang2017chestxray8}):} The network was trained on 112,120 chest radiographs. The proposed stochastic Poisson--Gaussian noise model generated Noise2Noise training image pairs.

    \item \textbf{Zero-shot Quantitative Evaluation (Montgomery County Dataset \cite{jaeger2014public}):} A total of 138 chest radiographs were used for zero-shot evaluation to assess the model's generalization capability.

    \item \textbf{Cross-domain Validation (Shenzhen Dataset \cite{candemir2014lung}):} Additional validation was performed on an unseen chest radiograph from the Shenzhen Hospital dataset to evaluate cross-domain generalization under a different imaging distribution.
\end{enumerate}

The Montgomery and Shenzhen datasets were used exclusively for evaluation and were not included during training, ensuring an unbiased assessment of zero-shot restoration capability and cross-domain generalization.

Image restoration quality was evaluated using two widely adopted performance metrics: the Peak Signal-to-Noise Ratio (PSNR), which measures pixel-level reconstruction accuracy, and the Structural Similarity Index Measure (SSIM), which evaluates structural similarity, local contrast, and anatomical detail preservation.

\subsection{Implementation Details}

The proposed \textit{Denoise X} U-Net architecture, comprising approximately 7.79 million trainable parameters, was implemented and evaluated using an NVIDIA Tesla P100 GPU. Gaussian, Median, and Bilateral filters were implemented using standard CPU-based libraries. To investigate deployment in resource-constrained environments, the trained TensorFlow model was converted into a lightweight 31.16\,MB ONNX model for CPU inference.

During inference, the reconstructed patches were seamlessly assembled to generate
high-resolution chest radiographs while preserving anatomical continuity across
patch boundaries.

\subsection{Quantitative Assessment}

\begin{figure*}[!t]
\centering

\begin{tabular}{cc}
\includegraphics[width=0.47\textwidth]{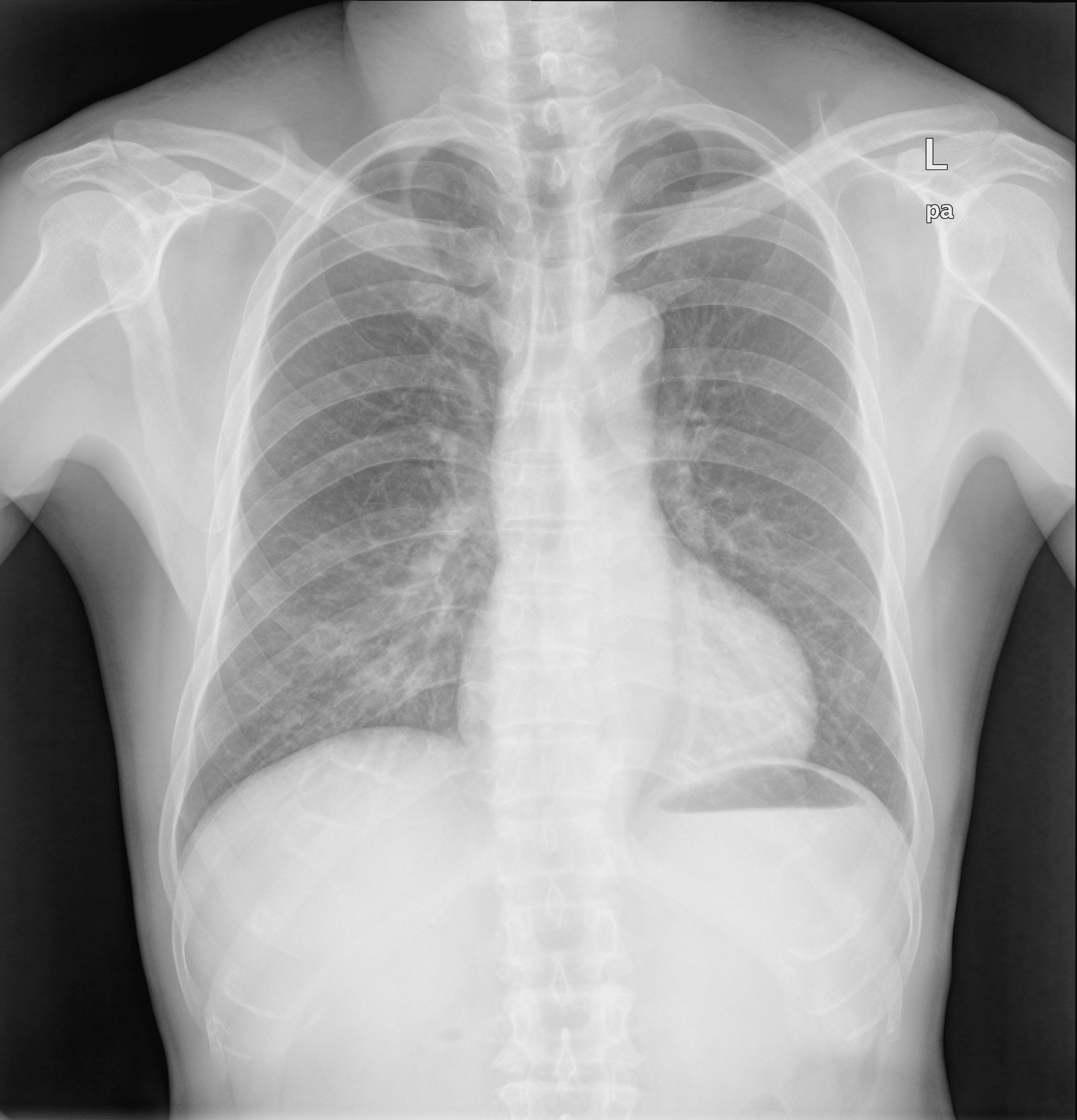} &
\includegraphics[width=0.47\textwidth]{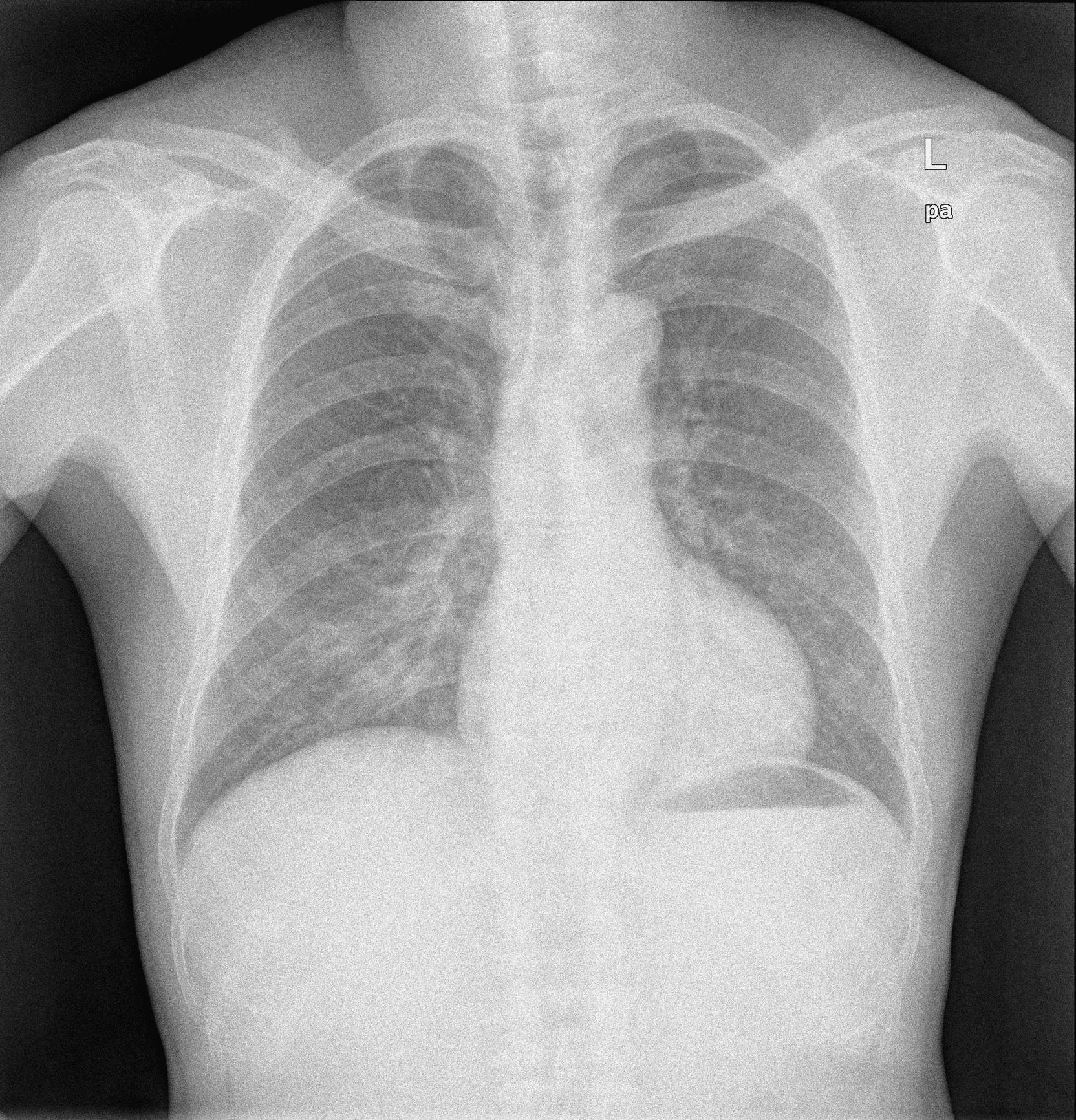} \\[2mm]

(a) Ground Truth &
(b) Noisy Input \\[4mm]

\includegraphics[width=0.47\textwidth]{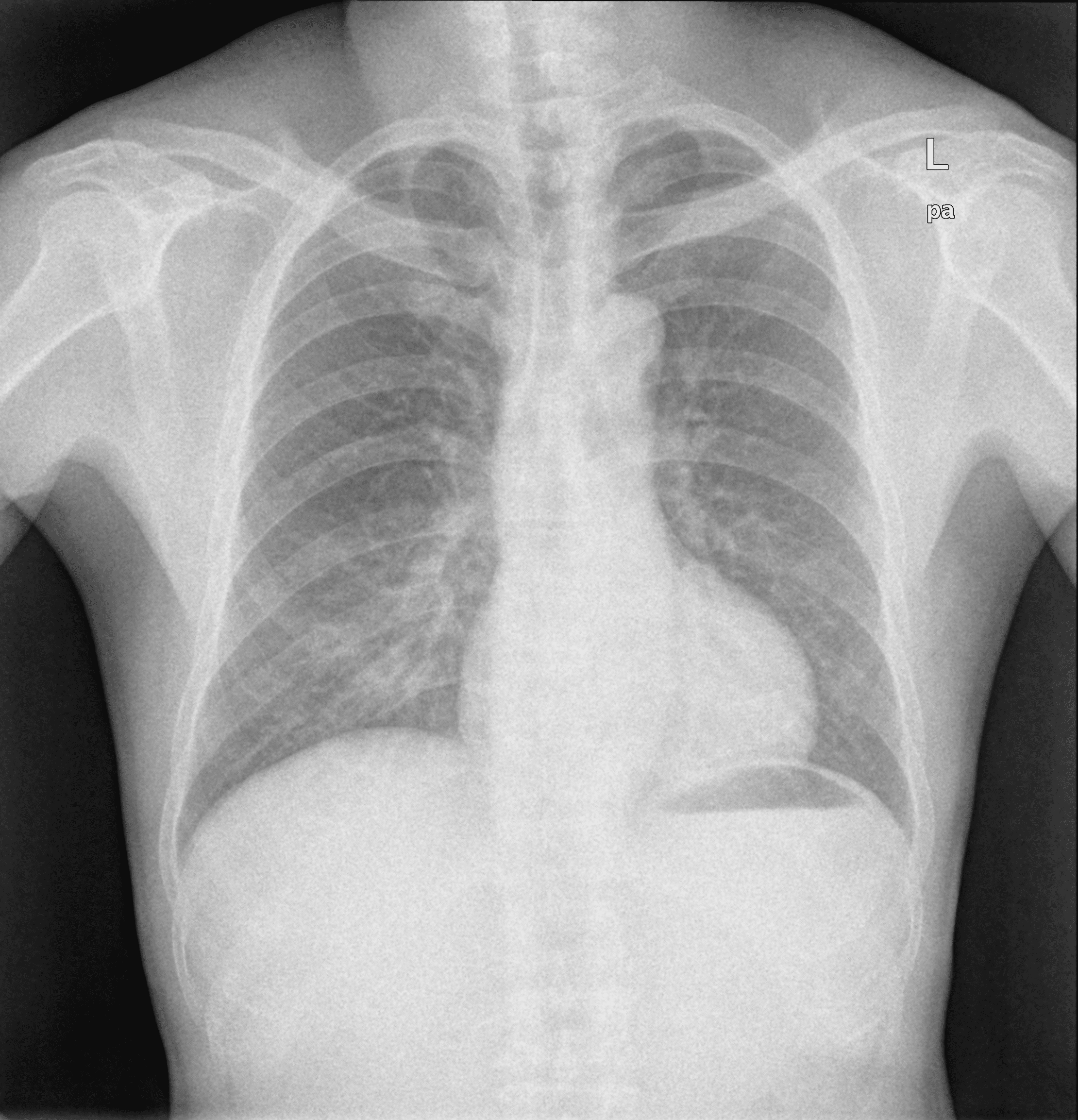} &
\includegraphics[width=0.47\textwidth]{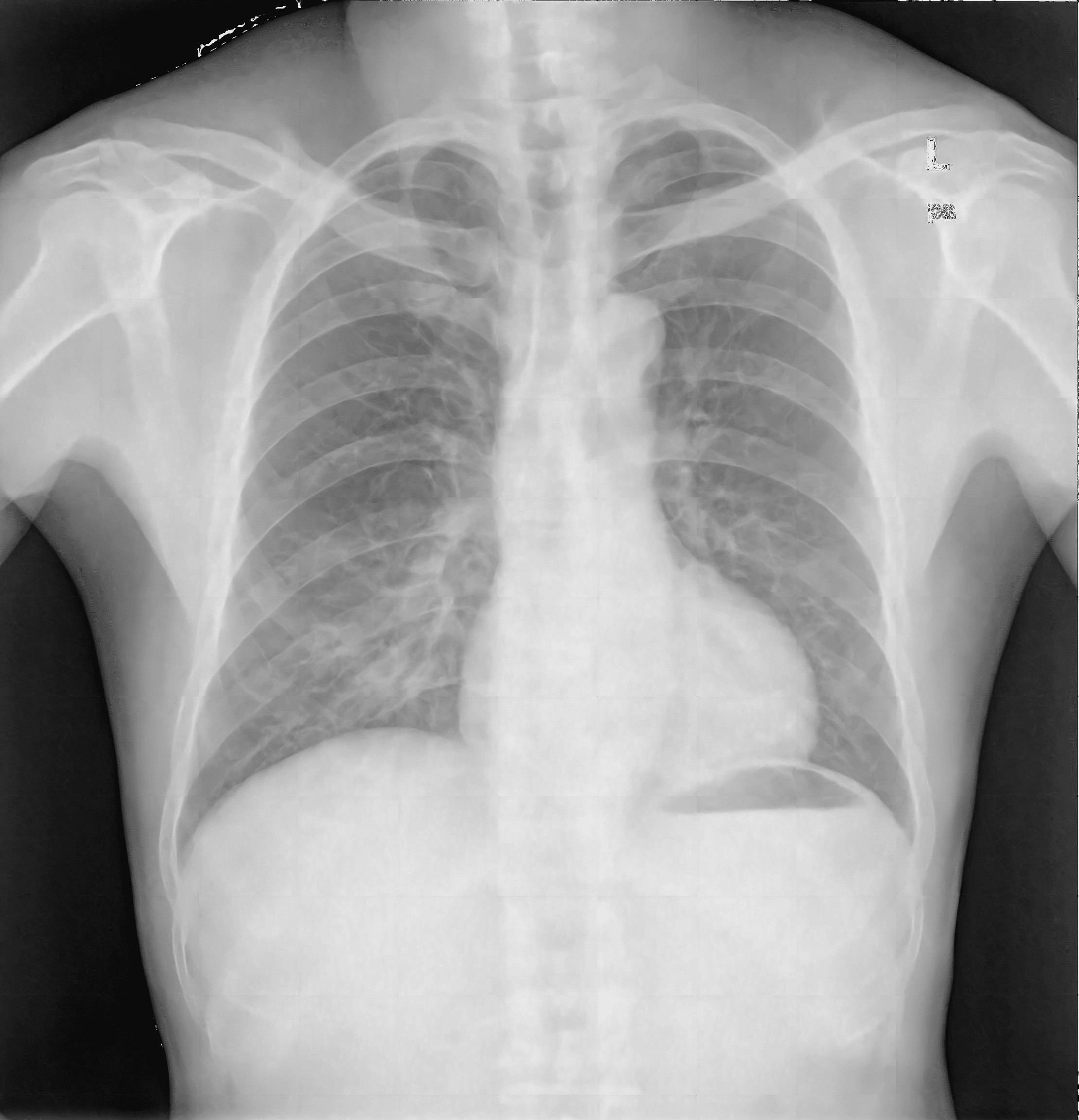} \\[2mm]

(c) Bilateral Filter &
(d) Denoise X
\end{tabular}

\caption{Qualitative comparison of restoration results
on an unseen chest radiograph from the Shenzhen
dataset~\cite{candemir2014lung}.}
\label{fig:qualitative_results}

\end{figure*}

Table~\ref{tab:montgomery_results} presents the comparative performance of the proposed method against classical filtering techniques on the unseen Montgomery County dataset~\cite{jaeger2014public}. The artificially degraded input images exhibited an average SSIM of 0.1300, indicating severe degradation caused by simulated quantum mottle. Among the classical filtering approaches, the Bilateral filter achieved the best baseline performance with a PSNR of 31.03~dB and an SSIM of 0.6431. However, because it relies on neighborhood statistics, it tends to smooth fine anatomical structures together with image noise, resulting in noticeable edge blurring, as shown in Fig.~\ref{fig:qualitative_results}(c)

The proposed \textit{Denoise X} framework outperformed all baseline methods, achieving a PSNR of 36.09~dB and an SSIM of 0.8806. This corresponds to an SSIM improvement of +0.2375 over the best baseline, demonstrating the effectiveness of the proposed self-supervised framework in suppressing Poisson--Gaussian noise while preserving diagnostically important anatomical structures, as shown in Fig.~\ref{fig:qualitative_results}(d). These quantitative results confirm that the proposed restoration framework achieves a favorable balance between effective noise suppression and preservation of diagnostically relevant image structures, consistently outperforming conventional filtering techniques on previously unseen chest radiographs.

\subsection{Generalization across Datasets}

For testing generalization performance outside the NIH ChestX-ray14~\cite{wang2017chestxray8} and Montgomery County~\cite{jaeger2014public} datasets, cross-domain testing was performed on an unseen chest radiograph from the Shenzhen Hospital dataset~\cite{candemir2014lung}. As shown in Table~\ref{tab:shenzhen_validation}, Denoise X yielded a PSNR of 36.44~dB and an SSIM of 0.9099, which were significantly better than those of the Bilateral filter, which achieved an SSIM of 0.7493. These results suggest that the proposed stochastic Poisson--Gaussian noise model facilitates the learning of generalized anatomical representations, even under varying clinical image distributions and acquisition systems.

\begin{table*}[!t]
\centering
\caption{Quantitative Performance Comparison on the Montgomery County Dataset (138 Unseen Chest Radiographs)}
\label{tab:montgomery_results}
\begin{tabular}{lcccc}
\toprule
\textbf{Restoration Method} & \textbf{PSNR (dB)} & \textbf{SSIM} & \textbf{Platform} & \textbf{Runtime} \\
\midrule
Simulated Noise (Baseline) & 20.02 & 0.1300 & N/A & N/A \\
Gaussian Blur & 29.13 & 0.5539 & CPU & 10.77 ms \\
Median Blur & 31.08 & 0.6644 & CPU & 27.72 ms \\
Bilateral Filter & 31.03 & 0.6431 & CPU & 510.35 ms \\
\textbf{Denoise X (GPU)} & \textbf{36.09} & \textbf{0.8806} & \textbf{Tesla P100 GPU} & \textbf{2.22 s} \\
\textbf{Denoise X (ONNX)} & \textbf{36.09} & \textbf{0.8806} & \textbf{CPU (ONNX Runtime)} & \textbf{$\sim$120 s} \\
\bottomrule
\end{tabular}
\end{table*}

\begin{table}[!t]
\centering
\caption{Cross-Domain Validation on an Unseen Shenzhen Chest Radiograph}
\label{tab:shenzhen_validation}
\begin{tabular}{lcc}
\toprule
\textbf{Method} & \textbf{PSNR (dB)} & \textbf{SSIM} \\
\midrule
Simulated Noise (Baseline) & 19.80 & 0.1357 \\
Bilateral Filter & 32.90 & 0.7493 \\
\textbf{Denoise X (Proposed)} & \textbf{36.44} & \textbf{0.9099} \\
\bottomrule
\end{tabular}
\end{table}

\begin{figure}[!t]
    \centering
    \includegraphics[width=\columnwidth]{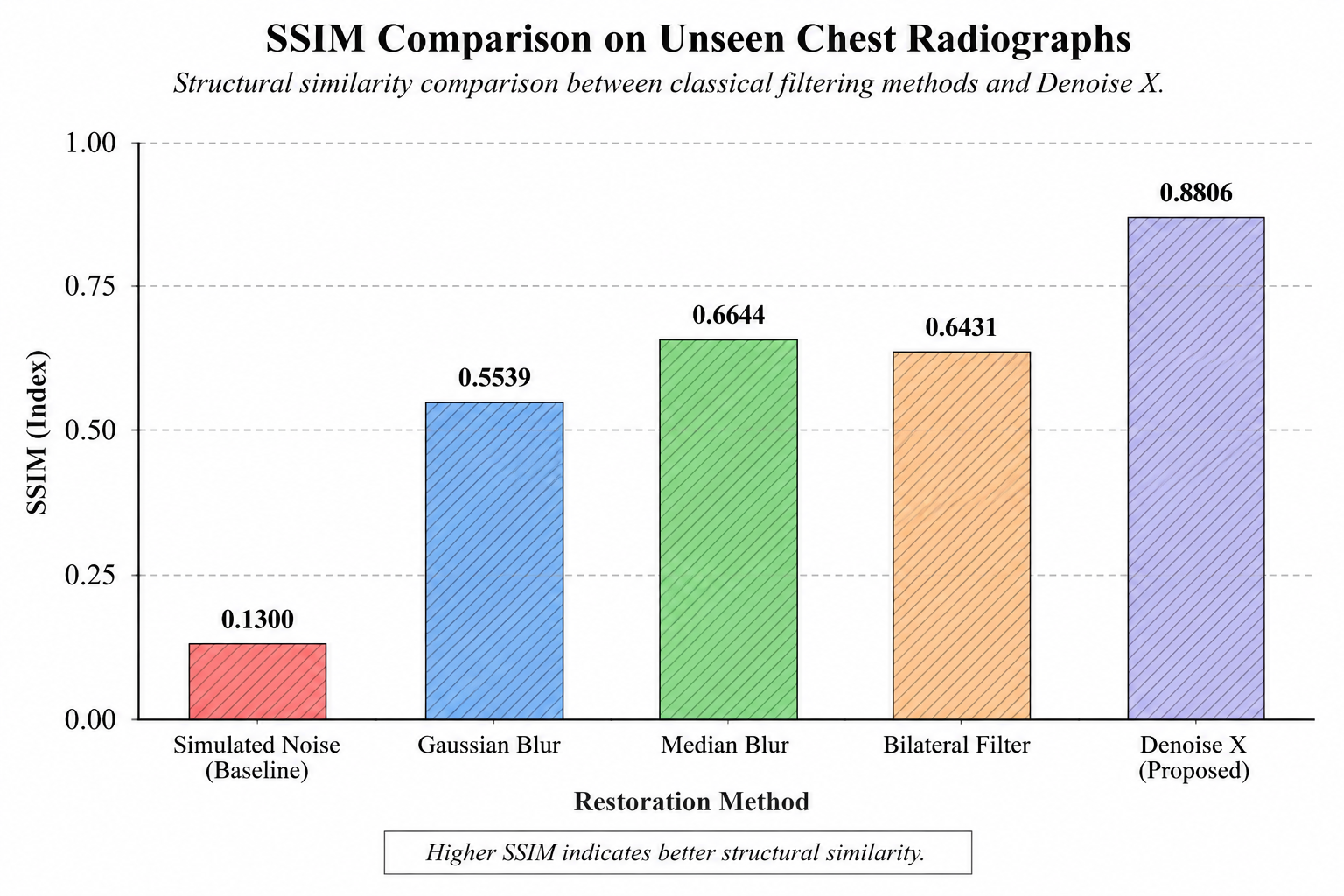}
    \caption{SSIM comparison of Denoise X with classical filtering methods on unseen chest radiographs.}
    \label{fig:ssim_comparison}
\end{figure}

\subsection{Computational Complexity and Deployment Analysis}

One of the main objectives of the proposed framework is to enable deployment in resource-constrained clinical settings.. Running inference on native high-resolution multi-megapixel chest radiographs using the proposed patch-based inference method takes roughly \textbf{2.2~s} per image on NVIDIA Tesla P100 GPU (Table~\ref{tab:montgomery_results}). Running the same native resolution inference method using ONNX model on a regular CPU will increase the inference time to about \textbf{120~s} per radiograph.

However, such latency becomes undesirable for mobile and point-of-care applications, which require optimizations.To address this limitation, an adaptive preprocessing stage was implemented to dynamically scale extreme-resolution images of the extreme-resolution images obtained from the mobile device in either a normal JPEG or PNG format, with a maximum of 2048~pixels while maintaining the aspect ratio intact. This helps to cut down the number of inference patches from about 192 to 48, thus reducing the time taken by ONNX CPU inference from about 120 seconds to merely \textbf{3--5~seconds}, all without using any GPU. Native DICOM medical images bypass this scaling step to preserve their raw, uncompressed bit-depth and structural fidelity.

\section{Results and Discussion}

The experimental results demonstrate the effectiveness of the proposed Denoise X architecture in bridging high-quality medical image restoration with the constraints of edge deployment. Using the self-supervised Noise2Noise approach, the network effectively suppresses quantum mottle and electronic readout noise without requiring paired ground-truth radiographs.

From a clinical perspective, preserving high-frequency anatomical structures is essential. Classical methods such as the Bilateral filter rely on local statistics and, although they reduce noise, can blur fine anatomical structures, as reflected by an SSIM of 0.6431 on the Montgomery County dataset~\cite{jaeger2014public}. In contrast, the proposed U-Net achieved an SSIM of 0.8806, indicating improved preservation of structural information, including pulmonary vasculature and bony structures. Furthermore, cross-dataset evaluation on the Shenzhen dataset~\cite{candemir2014lung}, with an SSIM of 0.9099, demonstrates promising generalization across different imaging distributions and acquisition systems through the proposed stochastic Poisson--Gaussian noise simulation.

The 31.16~MB ONNX patch-wise inference pipeline addresses a major challenge in deploying AI-based restoration systems in resource-constrained environments. While modern transformer architectures often require substantial computational resources, inference on native-resolution images using the ONNX model takes approximately 120~seconds per radiograph on a CPU, which may be suitable for asynchronous workflows. More importantly, the adaptive client-side preprocessing module downsizes high-resolution non-DICOM images to a maximum of 2048~px, reducing the number of inference patches and lowering CPU inference time to approximately 3--5~seconds.

\subsection{Limitations and Future Work}

While the proposed Denoise X framework demonstrates promising performance, several limitations remain:

\begin{enumerate}
    \item \textbf{Synthetic Noise Approximation:} The stochastic Poisson--Gaussian model closely approximates quantum mottle but cannot fully represent all scanner-specific and real-world clinical noise characteristics.

    \item \textbf{Clinical Validation:} Although PSNR and SSIM indicate strong restoration quality, future work will include radiologist-based Mean Opinion Score (MOS) studies to assess diagnostic usefulness.

    \item \textbf{Generalization:} Future research will evaluate the framework on images acquired from different imaging systems and clinical environments to improve robustness.

    \item \textbf{Model Enhancement:} Lightweight attention mechanisms and further optimization will be explored to improve structural preservation while maintaining efficient edge deployment.
\end{enumerate}

\section{Conclusion}

In this paper, Denoise X was introduced as a self-supervised, edge-deployable U-Net architecture for suppressing quantum mottle and electronic readout noise in low-dose chest radiographs. The integration of stochastic Poisson--Gaussian noise generation with the Noise2Noise learning paradigm enables training without requiring perfectly paired clean and noisy clinical radiographs.

Zero-shot evaluation on the unseen Montgomery County dataset~\cite{jaeger2014public} demonstrated that the 7.79-million-parameter network substantially outperformed the evaluated classical spatial filtering methods. Denoise X achieved a PSNR of 36.09~dB and an SSIM of 0.8806, compared with 31.03~dB and 0.6431, respectively, for the Bilateral filter, while preserving diagnostically important high-frequency anatomical structures. Furthermore, cross-domain evaluation on the unseen Shenzhen dataset~\cite{candemir2014lung} achieved a PSNR of 36.44~dB and an SSIM of 0.9099, indicating promising generalization across different imaging distributions.

In addition, the proposed framework addresses computational constraints associated with deployment in resource-limited clinical environments. By combining patch-based inference with conversion to a lightweight 31.16~MB ONNX model, Denoise X enables high-resolution radiograph restoration using CPU-based inference. These results demonstrate the potential of the proposed framework for practical edge deployment while providing a foundation for future validation using real low-dose clinical data and radiologist-based assessment.


\begin{thebibliography}{00}

\bibitem{tomasi1998bilateral}
C. Tomasi and R. Manduchi, ``Bilateral filtering for gray and color images,'' in \textit{Proc. IEEE Int. Conf. Comput. Vis. (ICCV)}, 1998, pp. 839--846.

\bibitem{zhang2017beyond}
K. Zhang, W. Zuo, Y. Chen, D. Meng, and L. Zhang, ``Beyond a Gaussian denoiser: Residual learning of deep CNN for image denoising,'' \textit{IEEE Trans. Image Process.}, vol. 26, no. 7, pp. 3142--3155, 2017.

\bibitem{ronneberger2015unet}
O. Ronneberger, P. Fischer, and T. Brox, ``U-Net: Convolutional networks for biomedical image segmentation,'' in \textit{Med. Image Comput. Comput.-Assist. Interv. (MICCAI)}, 2015, pp. 234--241.

\bibitem{lehtinen2018noise2noise}
J. Lehtinen, J. Munkberg, J. Hasselgren, S. Laine, T. Karras, M. Aittala, and T. Aila, ``Noise2Noise: Learning image restoration without clean data,'' in \textit{Proc. Int. Conf. Mach. Learn. (ICML)}, 2018, pp. 2965--2974.

\bibitem{krull2019noise2void}
A. Krull, T.-O. Buchholz, and F. Jug, ``Noise2Void---Learning denoising from single noisy images,'' in \textit{Proc. IEEE/CVF Conf. Comput. Vis. Pattern Recognit. (CVPR)}, 2019, pp. 2129--2137.

\bibitem{zamir2022restormer}
S. W. Zamir, A. Arora, S. Khan, M. Hayat, F. S. Khan, and M.-H. Yang, ``Restormer: Efficient transformer for high-resolution image restoration,'' in \textit{Proc. IEEE/CVF Conf. Comput. Vis. Pattern Recognit. (CVPR)}, 2022, pp. 5728--5739.

\bibitem{chen2022simple}
L. Chen, X. Chu, X. Zhang, and J. Sun, ``Simple baselines for image restoration,'' in \textit{Eur. Conf. Comput. Vis. (ECCV)}, 2022, pp. 17--33.

\bibitem{wang2017chestxray8}
X. Wang, Y. Peng, L. Lu, Z. Lu, M. Bagheri, and R. M. Summers, ``ChestX-ray8: Hospital-scale chest X-ray database and benchmarks on weakly-supervised classification and localization of common thorax diseases,'' in \textit{Proc. IEEE Conf. Comput. Vis. Pattern Recognit. (CVPR)}, 2017, pp. 2097--2106.

\bibitem{bushberg2011essential}
J. T. Bushberg, J. A. Seibert, E. M. Leidholdt, Jr., and J. M. Boone, \textit{The Essential Physics of Medical Imaging}, 3rd ed. Philadelphia, PA, USA: Lippincott Williams \& Wilkins, 2011.

\bibitem{foi2008practical}
A. Foi, M. Trimeche, V. Katkovnik, and K. Egiazarian, ``Practical Poissonian-Gaussian noise modeling and fitting for single-image raw-data,'' \textit{IEEE Trans. Image Process.}, vol. 17, no. 10, pp. 1737--1754, 2008.

\bibitem{jaeger2014public}
S. Jaeger, S. Candemir, S. Antani, Y.-X. J. Wáng, P.-X. Lu, and G. Thoma, ``Two public chest X-ray datasets for computer-aided screening of pulmonary diseases,'' \textit{Quant. Imaging Med. Surg.}, vol. 4, no. 6, pp. 475--480, 2014.

\bibitem{candemir2014lung}
S. Candemir, S. Jaeger, K. Palaniappan, J. P. Musco, R. K. Singh, Z. Xue, A. Karargyris, S. Antani, G. Thoma, and C. J. McDonald, ``Lung segmentation in chest radiographs using anatomical atlases with nonrigid registration,'' \textit{IEEE Trans. Med. Imaging}, vol. 33, no. 2, pp. 577--590, 2014.

\end{thebibliography}
\end{document}